%% file: main.tex
\documentclass[letterpaper, 10 pt, conference]{ieeeconf}

\bstctlcite{bstctl:etal, bstctl:nodash, bstctl:simpurl}

\IEEEoverridecommandlockouts                              

\usepackage{epsfig}
\usepackage{graphicx}
\usepackage{amsmath}
\usepackage{subcaption}
\usepackage{mathptmx} 
\usepackage{xcolor}
\usepackage{multirow}
\usepackage{pifont}

\newcommand{\cmark}{\ding{51}}%
\newcommand{\xmark}{\ding{55}}%

\definecolor{darkblue}{RGB}{0, 0, 200}

\makeatletter
\let\NAT@parse\undefined
\DeclareRobustCommand\onedot{\futurelet\@let@token\@onedot}
\def\@onedot{\ifx\@let@token.\else.\null\fi\xspace}

\usepackage[colorlinks,pagebackref=false,citecolor=blue,bookmarks=false,hypertexnames=true]{hyperref}

\title{Monocular Instance Motion Segmentation for 
Autonomous Driving: \\ KITTI InstanceMotSeg Dataset and Multi-task Baseline }

\author{Eslam Mohamed$^{*}$$^{1}$, Mahmoud Ewaisha$^{*}$$^{1}$, Mennatullah Siam$^{2}$, Hazem Rashed$^{1}$, \\
Senthil Yogamani$^{3}$, Waleed Hamdy$^{1}$, Mohamed  El-Dakdouky$^{4}$ and Ahmad El-Sallab$^{1}$ \\
{\normalsize 
$^{*}$Equal contribution \hspace{0.2cm} 
$^{1}$Valeo Egypt\hspace{0.2cm} 
$^{2}$University of Alberta \hspace{0.2cm} 
$^{3}$Valeo Ireland \hspace{0.2cm} 
$^{4}$Zewail City of Science and Technology
} 
}

\begin{document}

\bstctlcite{IEEEexample:BSTcontrol}

\maketitle

\begin{abstract}

\input{content/abstract}

\end{abstract}

\section{INTRODUCTION}
\label{sec:introduction}
\input{content/intro}

\section{Related Work}
\label{sec:related}
\input{content/related}

\section{Proposed Method} \label{sec:method}
\input{content/method}

\section{Experimental Results}
\input{content/exps}

\section{Conclusions}
\input{content/conc}

\section*{ACKNOWLEDGEMENTS}
We would like to thank  B Ravi Kiran (Navya), Letizia Mariotti and Lucie Yahiaoui for reviewing the paper and providing feedback.


\bibliographystyle{IEEEtran}
\bibliography{egbib1}


\end{document}

%% file: content/abstract.tex
Moving object segmentation is a crucial task for autonomous vehicles as it can be used to segment objects in a class agnostic manner based on their motion cues. It enables the detection of unseen objects during training (e.g., moose or a construction truck) based on their motion and independent of their appearance. Although pixel-wise motion segmentation has been studied in autonomous driving literature, it has been rarely addressed at the instance level, which would help separate connected segments of moving objects leading to better trajectory planning. 
As the main issue is the lack of large public datasets, we create a new InstanceMotSeg dataset comprising of 12.9K samples improving upon our KITTIMoSeg dataset. In addition to providing instance level annotations, we have added 4 additional classes which is crucial for studying class agnostic motion segmentation.
We adapt YOLACT and implement a motion-based class agnostic instance segmentation model which would act as a baseline for the dataset. We also extend it to an efficient multi-task model which additionally provides semantic instance segmentation sharing the encoder. The model then learns separate prototype coefficients within the class agnostic and semantic heads providing two independent paths of object detection for redundant safety.
To obtain real-time performance, we study different efficient encoders and obtain 39 fps on a Titan Xp GPU using MobileNetV2 with an improvement of 10\% mAP relative to the baseline. Our model improves the previous state of the art motion segmentation method by 3.3\%. The dataset and qualitative results video are shared in our website at \url{https://sites.google.com/view/instancemotseg}.

%% file: content/intro.tex

Motion cues play a significant role in automated driving. Most of the modern automated driving systems leverage HD maps to perceive the static infrastructure \cite{milz2018visual} while moving objects are more critical to be detected accurately. Due to the dynamic and interactive nature of moving objects, it is necessary to understand the motion model of each surrounding object to predict their future state and plan the ego-vehicle trajectory. Historically, the initial prototypes of automated driving relied on depth and motion estimation without any appearance based detection \cite{franke1998autonomous,kumar2018monocular,ravi2018real}. Therefore motion segmentation is an important perception task for autonomous driving~\cite{siam2018modnet,siam2018real,ramzy2019rst,Rashed_2019_ICCV_Workshops}. It can also to be used as an auxiliary task for end-to-end systems. Detection of moving objects can be seen as an alternate way to detecting objects and thus providing redundancy needed for a safe and robust system. In this case we call it class agnostic segmentation as it is solely dependant on motion cues regardless of semantics which would scale better to unknown objects. Since it is practically infeasible to collect data for every possible object category, this class agnostic segmentation can detect such unseen objects (e.g. rare animals like moose or kangaroo or rare vehicles like construction trucks). 

\begin{figure}[t!]
\centering
\includegraphics[width=.45\textwidth]{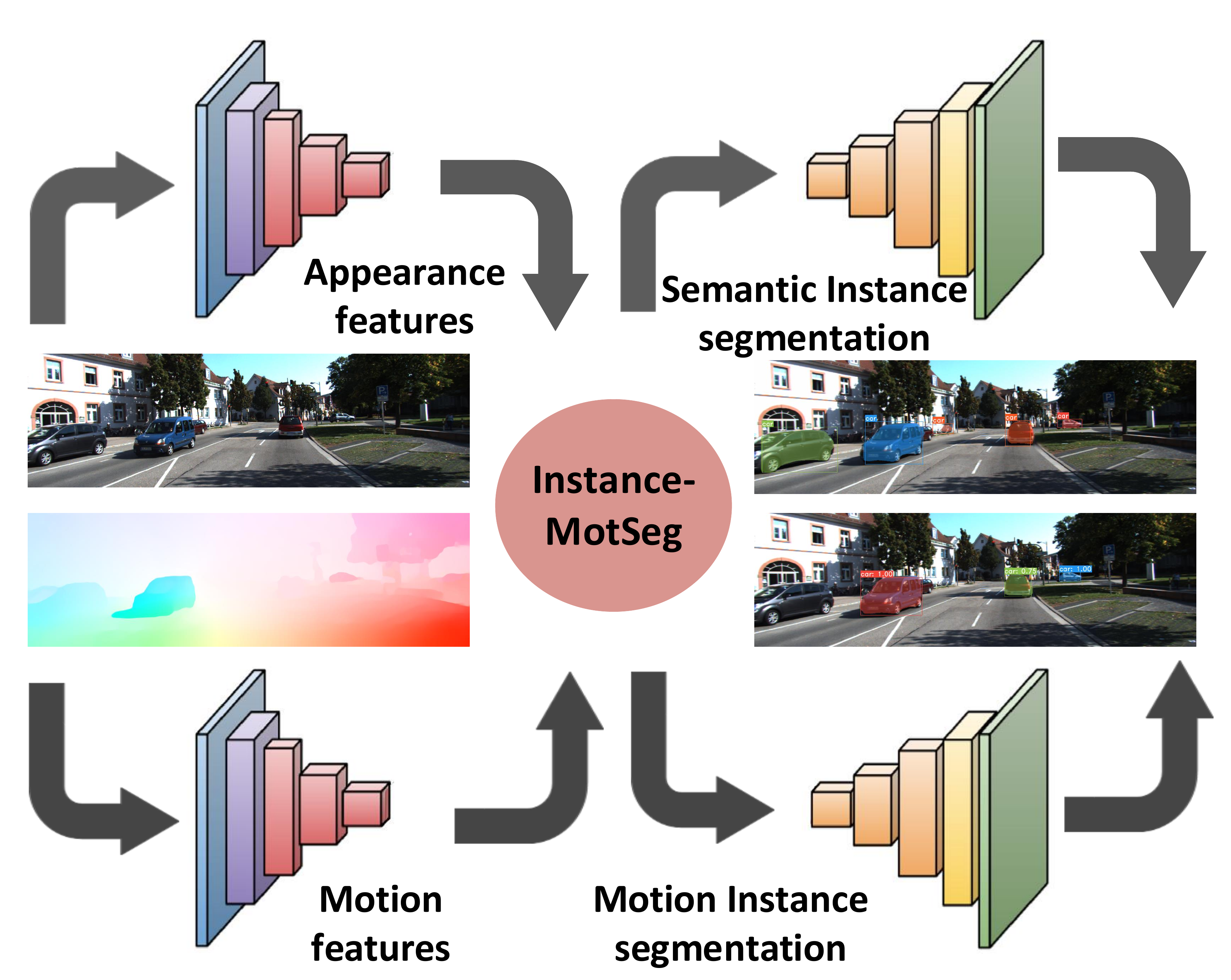}
\caption{Overview of the proposed network architecture.} 
\label{fig:overview}
\vspace{-0.4cm}
\end{figure}

Semantic segmentation and object detection are mature algorithms due to the large scale datasets available for these tasks \cite{siam2017deep}. Relatively, motion segmentation has limited datasets \cite{Valada_2017_IROS,siam2018modnet} and they are primarily focused on vehicles. Additionally, there are fundamental challenges in motion estimation which are not studied in detail: (1) Camera on moving vehicle makes it challenging to decouple ego-motion from motion of other moving objects. (2) Motion parallax and ambiguity in degenerate cases where objects are moving same direction as the ego-vehicle. (3) Limited computational power available in embedded platforms where an efficient CNN model for all tasks has to be designed \cite{briot2018analysis}. Pure geometric approaches were not able to fully solve the motion segmentation problem and most recent work~\cite{Valada_2017_IROS,siam2018real,Rashed_2019_ICCV_Workshops} focus either on learning based approaches solely or a hybrid of learning and geometric based approaches.




Semantic segmentation and motion segmentation do not separate instances of objects, a segment could comprise of several objects. In order to plan an optimal trajectory, it is necessary to extract individual objects. In case of semantics, this is handled by instance segmentation task which can be broadly categorized as one-stage or two-stage methods~\cite{he2017mask,li2017fully,bolya2019yolact}. One-stage methods such as YOLACT~\cite{bolya2019yolact} are computationally more efficient than two-stage ones. YOLACT~\cite{bolya2019yolact} formulates instance segmentation as learning prototypes which form a basis of a vector space and then learns the linear combination weights of these prototypes per instance. In this paper, we build upon this idea of using prototypes and adapt it for class agnostic instance segmentation using motion cues. We propose to jointly learn semantic and class agnostic (motion) instance segmentation using a shared backbone~\cite{sistu2019neurall, chennupati2019multinet++} and protonet (prototype generation branch) weights with  shallow prediction heads for each task. The prediction heads for both semantic and class agnostic segmentation are responsible for predicting bounding boxes, classes and prototype coefficients. The separation of the two tasks allows for training the class agnostic head on separate datasets that provide class agnostic annotations such as DAVIS~\cite{Caelles_arXiv_2019}. To the best of our knowledge, this is the first attempt to perform joint semantic and class agnostic instance segmentation that would create a step toward fail safe systems.  Figure~\ref{fig:overview} shows an overview of our method. 
To summarize, the contributions of this work include:
\begin{itemize}
    \item Release of a new InstanceMotSeg dataset with improved annotations, instance labels and additional classes.
    \item Demonstration of a good prototype of instance motion segmentation  trained using the proposed dataset. 
    \item A real-time multi-task learning model for semantic and class agnostic instance segmentation using motion. Our method relies on learning different prototype coefficients per task based on YOLACT~\cite{bolya2019yolact}.
    \item Ablation study of different backbones to find the optimal accuracy vs speed trade-off and different types of architectures for encoding motion.
\end{itemize}

%% file: content/related.tex
\subsection{Motion Segmentation}

Motion analysis approaches can be roughly categorized into geometry, tracking or learning based methods. Geometry based approaches are typically pixel-based constrained with some global regularization and they typically do not exploit global context. It has been extensively studied in ~\cite{torr1998geometric,kundu2009moving,menze2015object,scott2017motion,bideau2016s}. Scott et. al.~\cite{scott2017motion} relied on a homography based method but it is simplistic for handling complex motion in automotive scenes. Bideau et. al.~\cite{bideau2016s} utilized optical flow and relied on perspective geometry and feature matching to perform motion segmentation. Tracking based methods~\cite{lin2014deep,brox2010object,ochs2011object} rely on computing object trajectories and can perform clustering to perform the final motion segmentation. These methods are computationally expensive. 

Most of the current progress is focused on learning based methods~\cite{tokmakov2016learning,tokmakov2017learning,Valada_2017_IROS,siam2018rtseg}. Tokmakov et al. initially explored optical flow based single stream fully convolutional network \cite{tokmakov2016learning} and then extended it to use both motion and appearance by learning a bidirectional recurrent model with memory~\cite{tokmakov2017learning}. Vertens et. al.~\cite{Valada_2017_IROS} proposed a method based on FlowNet~\cite{dosovitskiy2015flownet,Ilg2016FlowNet2E} to jointly learn semantic segmentation and motion segmentation. The model was trained on their dataset CityMotion based on Cityscapes~\cite{Cordts2016cityscapes}. Concurrently Siam et. al.~\cite{siam2018modnet} proposed a motion and appearance based multitask learning framework called MODNet to perform object detection and motion segmentation. MODNet was trained on KITTIMoseg that they released based on KITTI dataset~\cite{Geiger2013IJRR}. It was further extended using ShuffleNet encoder for computational efficiency~\cite{siam2018rtseg}. Rashed et. al. released a larger scale KITTIMoseg and experimented on using LIDAR flow as another mean to improve motion segmentation~\cite{Rashed_2019_ICCV_Workshops}. A similar approach to~\cite{siam2018modnet,rashed2019motion} that can generate motion segmentation weakly supervised annotations was used on WoodScape fisheye dataset~\cite{yogamani2019woodscape}. 

\textbf{Instance Motion Segmentation:} Segmentation of Independently Moving Objects is a well studied problem in classical computer vision literature \cite{bideau2020motion}. However, it has been relatively less explored in the area of autonomous driving due to lack of large datasets. In a rare attempt in \cite{cao2019learning}, instance motion segmentation was explored for stereo cameras. They reported that there was no prior work in this area. The results were reported on a small KITTI Sceneflow dataset containing 200 test images. Thus we were motivated to create a large scale dataset for instance level motion segmentation and implement a new baseline for monocular setting which is more challenging.



\begin{table*}[!t]
\caption{Comparison of different datasets for motion segmentation. } 
\label{table:datasets_motion}
\centering
\begin{tabular}{|l|l|l|l|l|}
\hline
 Datasets & \# Frames & \# Sequences & \# Object Categories & Instance Labels \\ \hline
 Cityscapes-Motion~\cite{Valada_2017_IROS} & 3475 & - & 1 (Car Only) & \xmark \\ \hline
 KITTI-SceneFlow~\cite{menze2015object} & 400 & - & 1 (Car Only) & \xmark \\ \hline
  KITTI-Motion~\cite{Valada_2017_IROS} & 455 & - & 1 (Car Only) & \xmark \\ \hline
 KITTI-MoSeg~\cite{siam2018modnet} & 1300 & $\sim$ 5 & 1 (Car Only) & \xmark \\ \hline
 KITTI-MoSeg Extended~\cite{Rashed_2019_ICCV_Workshops} & 12919 & $\sim$ 38 & 1 (Car Only) & \xmark \\ \hline
 KITTI-InstanceMotSeg (Ours) & 12919 & $\sim$ 38 & 5 & \cmark \\ \hline 
\end{tabular}

\end{table*}

\begin{table}[]
\caption{ Class distribution of moving and static objects in our dataset.
}
\label{tab:dataset_stats}
\centering
\resizebox{0.45\textwidth}{!}{
\begin{tabular}{|l|l|c|c|c|c|c|}
\hline
Type/Class & Car & Truck & Van  & Pedestrian & Cyclist  \\ \hline
Static & 29656 & 542 & 3388 & 927 & 243 \\
Moving & 8788 & 1117 & 1382 & 1358 & 1292\\
Total & 38444 & 1659 & 5220 & 2285 & 1535\\
\hline
\end{tabular}
}
\end{table}

\subsection{Video Object Segmentation (VOS)}

A related task which has been extensively studied in the literature on the DAVIS benchmark~\cite{Perazzi2016} is Video object segmentation. It can be categorized into semi-supervised and unsupervised approaches. Semi-supervised methods utilize a mask initialization to track the objects of interest through the video sequence~\cite{voigtlaender2017online,khoreva2016learning}. Unsupervised approaches do not require any initialization mask and they segment the primary object in the sequence based on appearance and motion saliency~\cite{kohprimary,tokmakov2017learning,jain2017fusionseg}. Most of the initial work in video object segmentation did not predict at instance level. The first attempts
to perform instance level video object segmentation was proposed by Hu et. al.~\cite{hu2017maskrnn} relying on recurrent neural networks. VOS setting is based on one large moving object in a scene which is the most salient. Relatively, automotive scenes are much more complex with multiple moving objects. The focus of this work is on automotive scenes. However, we use DAVIS'17 training data in some experiments to train our class agnostic head which provides higher variability in the classes presented in the dataset.


%% file: content/method.tex
\subsection{KITTI InstanceMotSeg Dataset}
\label{sec:instmoseg}


Table~\ref{table:datasets_motion} shows a comparison of the different datasets with respect to our KITTI InstanceMotSeg dataset. 
The main motivation of this dataset to create instance level motion segmentation annotations for multiple classes. None of the other motion segmentation datasets in the table provide instance level labels. However, it is easy to obtain for Cityscapes-Motion \cite{Valada_2017_IROS} using instance segmentation annotation provided in Cityscapes\cite{Cordts2016cityscapes}. Compared to Cityscapes-Motion, our dataset provides nearly 4x more samples and 5 classes instead of one. The other KITTI motion segmentation datasets namely KITTI-SceneFlow~\cite{menze2015object}, KITTI-Motion~\cite{Valada_2017_IROS} and KITTI-MoSeg~\cite{siam2018modnet} do not provide instance labels and have far lesser samples.

In our previous work, we created KITTI-MoSeg Extended~\cite{Rashed_2019_ICCV_Workshops} which had a significantly higher number of 12.9k samples. It made use of a semi-automated approach to obtain motion segmentation labels via finding difference in ego-motion and the motion of other objects in Velodyne coordinate system. This method produced erroneous annotations when the camera undergoes rotation due to the car  turning. In this work, we modified the annotation tool to estimate motion in 3D world coordinate system instead of the Velodyne coordinate system which significantly improved the accuracy of annotations.

We also extend the dataset with instance segmentation masks for 5 classes including car, pedestrian, bicycle, truck, van instead of the original annotations only for car class. Addition of more classes provides larger variability in appearance and thus improves the appearance agnostic detection of motion segmentation.
Our dataset comprises of 12.9k frames and we divide it to 80-20\% ratio for training and testing. Table \ref{tab:dataset_stats} demonstrates the class distribution in our  dataset for both moving and static states.


\subsection{Instance-Level Class Agnostic Segmentation}
Our baseline model from YOLACT~\cite{bolya2019yolact} has a feature extractor network, a feature pyramid network~\cite{lin2017feature} and a protonet module that learns to generate prototypes. We experiment with different feature extractor networks as shown in section~\ref{sec:exps}. In parallel to protonet, a prediction head is learned that outputs bounding boxes, classes and coefficients that are linearly combined with the prototypes to predict the instance masks. We train this model for the class agnostic (motion) instance segmentation task to serve as a baseline for comparative purposes. The model is originally trained for semantic instance segmentation discarding the images with no mask annotations. On the other hand, in InstanceMotSeg dataset we use, 
there is a significant portion of frames that do not have moving objects. 
Utilization of such frames will provide a lot of negative samples of static cars which will help the model understand the appearance of static vehicles as well and thus reduce overfitting. In order to solve this problem, the loss function has been updated to reduce the confidence score of the predicted object in case of a false positive. The mask segmentation loss is formulated as binary cross entropy, while the bounding box regression uses a smooth L1 loss similar to SSD~\cite{liu2016ssd}. We refer to this baseline model \textit{RGB-Only} since the input modality is only appearance information.

   \begin{figure*}[t]
       \centering
 \includegraphics[width=.9\textwidth]{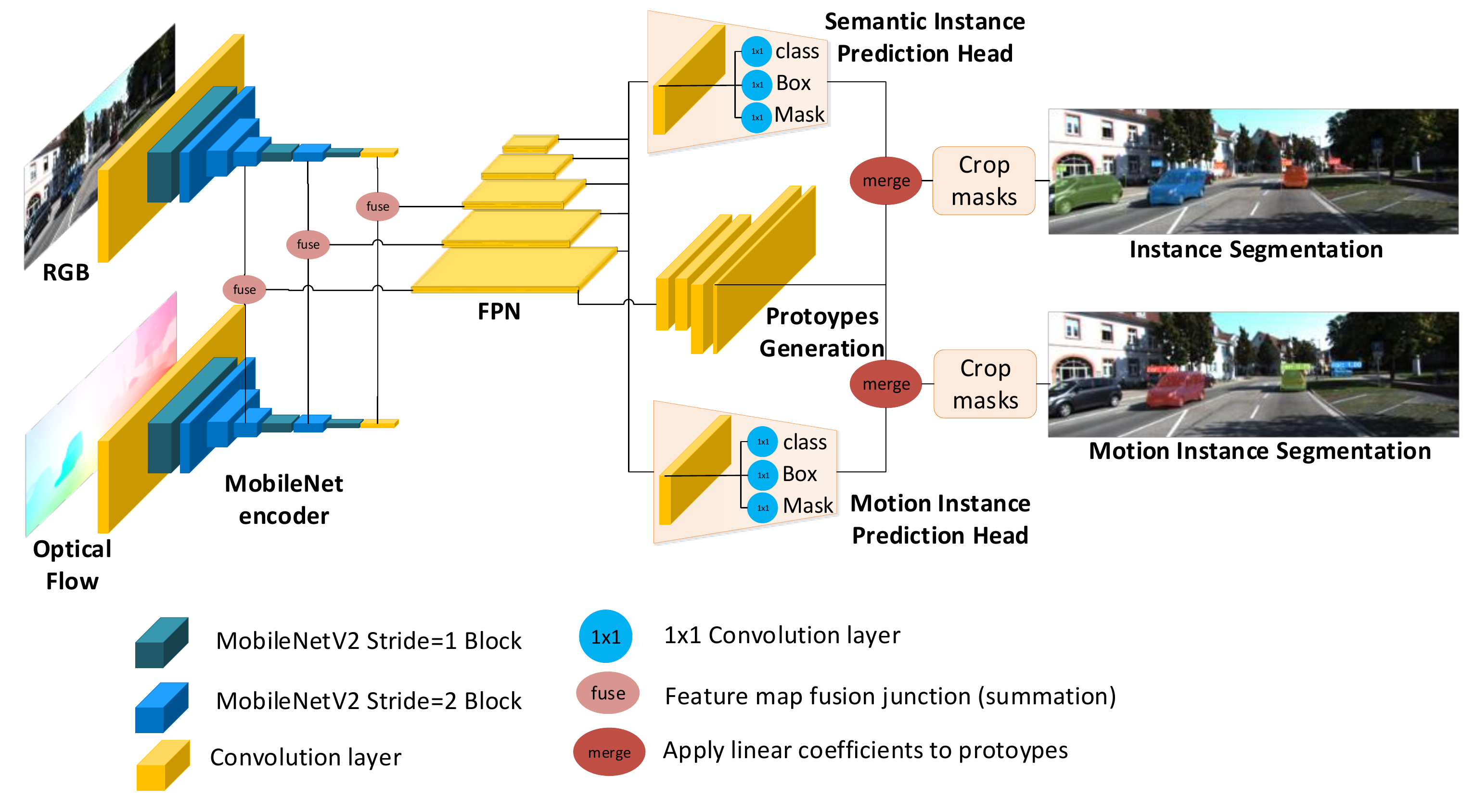}
     \caption{Proposed multi-task network architecture. RGB and optical flow  are fed as inputs to the network. Mask coefficients are learnt for two tasks separately using two prediction heads, namely semantic instance segmentation and motion instance segmentation. The coefficients of both prediction heads are combined with the prototypes generating semantic and motion instance masks. Our model has a real-time performance of 39 fps using MobileNet-V2 encoder. } 
     \label{fig:networkarch}
     \vspace{-0.4cm}
   \end{figure*}

We feed the model with motion information using two approaches and we provide a comparative study for their impact on motion instance segmentation. (1) Implicit method in which the network accepts a stack of images (t, t+1) and learns to model the motion implicitly as referred to by~\cite{ramzy2019rst}. (2) Explicit method where we use optical flow map highlighting pixels motion. In this approach, we make use of FlowNet 2.0~\cite{Ilg2016FlowNet2E} model to compute optical flow. Our main objective is to predict instance masks for moving objects. Hence, a motion representation, in our case optical flow, should be fed into the network in addition to appearance. As motion cues are important in automated driving, the standard automotive embedded platforms including Nvidia Xavier, TI TDA4x and Renesas V3H have a hardware accelerator for computing dense optical flow and it can be leveraged without requiring additional processing. Thus our runtime computation does not include optical flow processing step. Feature fusion method is adopted in our experiments in which we construct two separate feature extractors to process appearance and motion separately. The fusion is performed on the feature level which has provided significant improvement in~\cite{rashed2019motion, 8917447, Rashed_2019_ICCV_Workshops}. We refer to this feature-fusion model as \textit{RGB+OF}. The same architecture has been used to evaluate the impact of implicit motion modelling in which two sequential RGB images \textit{(t, t+1)} are input to the two-stream feature extractor. This model is referred to as \textit{RGB+RGB}.

\subsection{Multi-task Learning of Semantics and Class Agnostic Instances}
We further propose the joint model to learn semantic and class agnostic instance segmentation. The motivation is to provide a more generic representation for the surrounding environment including both static and moving instances while both tasks are learnt jointly. Separate prediction heads $D_s$ and $D_m$ are used for semantic and motion instance segmentation tasks respectively as shown in Figure~\ref{fig:networkarch}. 

During training, we alternate between $k$ steps for training $D_s$ semantic labels, and $k$ steps for training $D_m$ using InstanceMotSeg motion labels or the generic DAVIS dataset. One other approach to tackle the problem of learning both semantic and motion instance segmentation would be to pose it as a multi-label classification problem. However, to allow for the flexibility for training the model from different datasets that do not necessarily have the joint annotations and to allow for training motion instance segmentation as a class agnostic segmentation we follow our proposed approach. The separate motion instance segmentation head would be responsible for segmenting generally moving objects even if they are unknown for the semantic head. The feature extraction network, feature pyramid network and protonet are shared among the two tasks to ensure computational efficiency of our proposed multi-task learning system. Thus the learned prototypes are the same for both tasks but the learned coefficients to construct masks and the learned box regression are different.

For improving the functional safety of the system, the two outputs are independent and will be fused at the end in a sensor fusion module. For standard objects like vehicles, there will be two independent sources of detection based on motion and appearance. For objects unseen during training or not labelled explicitly, class agnostic motion instance segmentation will be used to detect.

%% file: content/exps.tex
\label{sec:exps}

\begin{table*}[t]
\caption{Quantitative comparison between different network architectures for class agnostic (motion) instance segmentation.}
\label{tab:quantitative}
\centering
\resizebox{0.8\textwidth}{!}{
\begin{tabular}{|l|l|c|c|c|c|c|c|c|c|c|}
\hline
\multirow{2}{*}{Model} & \multirow{2}{*}{Backbone} & \multirow{2}{*}{FPS} & \multirow{2}{*}{\#Params (M)} & \multirow{2}{*}{Time} & \multicolumn{3}{c|}{Mask} & \multicolumn{3}{c|}{Box} \\ \cline{6-11}
 &  &  & &  & AP  &   $\text{AP}_{50}$        & $\text{AP}_{75}$  &  AP     &      $\text{AP}_{50}$  & $\text{AP}_{75}$\\ \hline
\multicolumn{11}{|c|}{\textbf{[A] Comparison of Different Models}} \\ \hline
RGB-Only & ResNet101 & 34.71 & 49.6 & 47.9  &    28.38        &    40.88      &  33.93         &  30.12   & 41.67 &  36.65    \\ \hline
RGB+RGB(n-1) & ResNet101 & 20.3 & 95.9 & 32.05 &    31.2       &    32.05       &    35.55       & 31.2    & 48.94 &   36   \\ \hline
RGB+OF & ResNet101 & 20.3 & 95.9 & 49.2 &     39.26       &   60.02       &    45.76       &  41.24    & 60.59 & 49.28    \\ \hline
\multicolumn{11}{|c|}{\textbf{[B] Comparison of Different Backbones}} \\ \hline
RGB+OF & WS-ResNet101 & 20.54 & 49.6 & 48.6 &     43.1      &  62.3         &  48.5         & 43.6   & 61.86 &  53.26     \\ \hline
RGB+OF & EfficientNet & 35.06 & 21.2 & 28.5 &    40.41        &     59.56     &   43.05        & 38.57   & 59.38 & 44.43      \\ \hline
RGB+OF & MobileNetV2 & \textbf{39.18}  & 12.9 & \textbf{25.5} &     \textbf{43.55}      &    68.34       &   \textbf{48.59}        &   43.93    & 67.73 &  51.56  \\ \hline
RGB+OF & ShuffleNetV2 & 38 & \textbf{11.1} & 26.3 &     43.9      &   \textbf{68.43}        &    48.49       & \textbf{44.9}    & \textbf{68.26} &  \textbf{51.77}    \\ \hline
\end{tabular}
}
\end{table*}

\begin{table*}[t]
\caption{Quantitative Comparison between multi-task joint semantic and class agnostic instance segmentation against the baseline single tasks Baseline-S (semantic instance segmentation only) and Baseline-M (class agnostic (motion) instance segmentation only). 
}
\label{tab:quantitative_MTL}
\centering
\resizebox{\textwidth}{!}{
\begin{tabular}{|l|l|c|c|c|c|c|c|c|c|c|c|c|c|}
\hline
\multirow{3}{*}{Model} & \multirow{3}{*}{Backbone} & \multicolumn{6}{c|}{Semantic} & \multicolumn{6}{c|}{Class Agnostic}\\ \cline{3-14}
 & & \multicolumn{3}{c|}{Mask} & \multicolumn{3}{c|}{Box} & \multicolumn{3}{c|}{Mask} & \multicolumn{3}{c|}{Box}\\ \cline{3-14}
 & & AP  & $\text{AP}_{50}$ & $\text{AP}_{75}$ & AP & $\text{AP}_{50}$ & $\text{AP}_{75}$ & AP & $\text{AP}_{50}$ & $\text{AP}_{75}$ &  AP & $\text{AP}_{50}$ & $\text{AP}_{75}$ \\ \hline
 Baseline-M & ResNet101 & - & - & - & - & - & - & 39.26 & 60.02 & 45.76&  41.24 & 60.59 & 49.28 \\ 
 Baseline-S & ResNet101 & \textbf{28.7} & \textbf{50.3} & \textbf{27.4} & \textbf{32.5} & \textbf{61.6} & \textbf{30.4} & - & - & - &  - & - & - \\ 
 Multi-Task & ResNet101 & 22.5 & 39.5 & 22.2 & 27.7 & 54.5 & 25.7 & 44.3 & 66.7 & 50.4 & 44.8 & 66.8 & 55.1 \\ \hline
 Baseline-M & MobileNetV2 & - & - & - & - & - & - &  43.55 & 68.34 & 48.59 & 43.93 & 67.73 &  51.56   \\ 
 Baseline-S & MobileNetV2 & 25.9 & 47.0 & 25.0 & 30.0 & 58.3 & 53.0 & - & - & - &  - & - & - \\ 
 Multi-Task & MobileNetV2 & 21.9 & 36.5 & 22.6 & 26.5 & 53.6 & 24.2 & \textbf{45.3} & \textbf{69.0} & \textbf{57.4} & \textbf{44.3} & \textbf{69.4} & \textbf{52.4} \\ \hline
\end{tabular}
}
\end{table*}

In this section, we provide the details of our experimental setup and results on InstanceMotSeg dataset. Our model provides 9 fps speedup over other motion segmentation methods while additionally providing instance-wise motion masks.

\subsection{Experimental Setup}

We train our model using SGD with momentum using an initial learning rate of $10^{-4}$, momentum of 0.9 and a weight decay of $5x10^{-4}$. A learning rate scheduling is used where it is divided by 10 at iterations 280k and 600k. We train with batch size 4 for 150 number of epochs using ImageNet~\cite{denglarge} pretrained weights. The rest of the hyper-parameters and data augmentation technique are similar to YOLACT~\cite{bolya2019yolact}. For the dataset, we extend KITTIMoSeg with instance motion masks as explained in section~\ref{sec:instmoseg} to create InstanceMotSeg dataset which is used in training our network in addition to DAVIS'17~\cite{Caelles_arXiv_2019}. We plan to release the augmented labels for KITTIMoSeg to benefit researchers working on motion instance segmentation for autonomous driving. Finally, we report frame rate and time in milliseconds for all models running on Titan Xp GPU on image resolution 550 $\times$ 550 including FuseMODNet~\cite{Rashed_2019_ICCV_Workshops}.

\subsection{Benchmarking and Analysis}

\begin{table}[]
\caption{Quantitative comparison between our proposed model and state-of-the-art motion segmentation methods.
}
\label{tab:compare_motion_seg}
\centering
\resizebox{0.45\textwidth}{!}{
\begin{tabular}{|l|l|c|c|c|c|}
\hline
Model & Moving IoU & Background IoU & mIoU & FPS \\ \hline
RTMotSeg~\cite{siam2018real} & 50 & 99.1 & 74.6 & {25} \\
FuseMODNet~\cite{Rashed_2019_ICCV_Workshops} & 53.2 & 99.3 & 76.2 & 18 \\ 
Ours & \textbf{59.7} & \textbf{99.4} & \textbf{79.5} & \textbf{39} \\ \hline
\end{tabular}
}
\end{table}

We benchmarked our model with other state of the art motion segmentation methods in Table~\ref{tab:compare_motion_seg}. Since we are the first to propose instance motion segmentation in autonomous driving literature we post-process the instance motion masks into pixel-level motion segmentation and compare using mean intersection over union for moving pixels and frame rate. The closest work in motion segmentation literature is RTMotSeg~\cite{siam2018rtseg} and FuseMODNet~\cite{Rashed_2019_ICCV_Workshops}. Our model outperforms these methods in terms of mean IoU with 3.3\% and 4.9\% respectively. Also in terms of frames per second, our MobileNetV2 model is a more efficient solution with $\sim$ 39 fps than FuseMODNet model.

Furthermore, we provide exhaustive ablation studies in Table~\ref{tab:quantitative}. Two main factors are studied namely:
 (A) The impact of different input modalities to the model. In this study, we provide results which help us evaluate the importance of using optical flow against stack of images where the network learns to model motion implicitly. (B) The impact of various backbones to estimate the optimal performance in terms of accuracy vs speed trade-off. For this purpose, we explore ResNet101~\cite{he2016deep},
 EfficientNet~\cite{tan2019efficientnet}, ShuffleNetV2~\cite{ma2018shufflenet} and MobileNetV2~\cite{sandler2018mobilenetv2} as backbone architectures.

Table~\ref{tab:quantitative} demonstrates quantitative results evaluated on our dataset using various network architectures. Significant improvement of 11\% in mean average precision has been observed with feature-fusion which confirms the conclusions of \cite{siam2018modnet,yahiaoui2019fisheyemodnet,rashed2019motion}. Due to limitations of embedded devices used in autonomous driving, we focus our experiments on achieving the best balance between accuracy and speed using various network architectures. MobileNetV2 has the best run-time while maintaining the second best mean average precision. The best performing backbone is ShufflenetV2 with an insignificant improvement over MobileNetV2 with 0.3\%, but lower frames per second. {To our surprise, MobileNetV2 provides higher accuracy than the more complex ResNet101 architecture. It is probably because of the overfit to the training data and a higher network capacity is not required for the proposed task on InstanceMotSeg dataset.} Thus, MobileNetV2 provides the best speed-accuracy trade-off. 

\begin{figure*}[t]
\captionsetup[subfigure]{labelformat=empty}
\centering
\begin{subfigure}{.498\textwidth}
    \includegraphics[width=\textwidth]{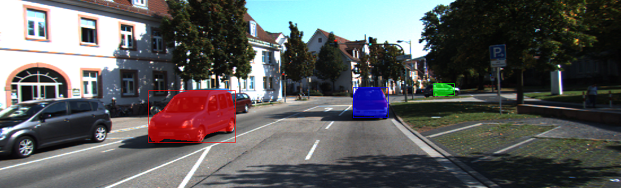}
    \vspace{-1cm}
    \caption{\textcolor{white}{(a)}}
\end{subfigure}%
\hfill
\begin{subfigure}{.498\textwidth}
    \includegraphics[width=\textwidth]{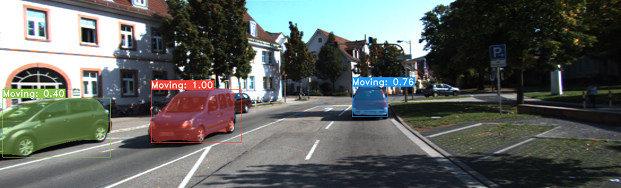}
    \vspace{-1cm}
    \caption{\textcolor{white}{(b)}}
\end{subfigure}%
\quad

\begin{subfigure}{.498\textwidth}
    \includegraphics[width=\textwidth]{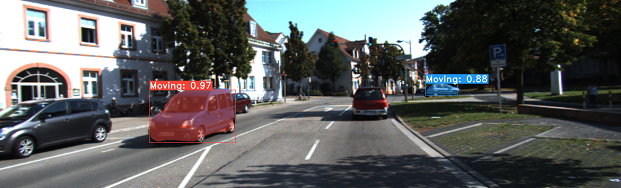}
    \vspace{-1cm}
    \caption{\textcolor{white}{(c)}}
\end{subfigure}%
\hfill
\begin{subfigure}{.498\textwidth}
    \includegraphics[width=\textwidth]{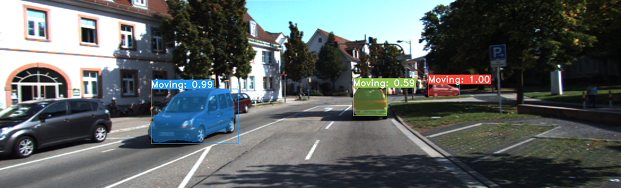}
    \vspace{-1cm}
    \caption{\textcolor{white}{(d)}}
\end{subfigure}
\quad
\caption{Qualitative results using our model on KITTI dataset. \textbf{(a)} Ground Truth, \textbf{(b)} RGB-only, \textbf{(c)} Feature-Fusion EfficientNet, \textbf{(d)} Feature-Fusion MobileNet}
\label{fig:qualitativeEval_Single}
\end{figure*}

Figure \ref{fig:qualitativeEval_Single} shows our qualitative results on KITTI dataset. (a) shows the ground truth of the moving instances using weak annotations in \cite{Rashed_2019_ICCV_Workshops}. (b) demonstrates inability of the baseline network to distinguish between moving and static objects using only color images without feeding the network with any motion signal.
(c), (d) demonstrate our results using different backbones where MobileNetV2 show the best accuracy vs speed trade-off as shown in Table \ref{tab:quantitative}.

\begin{table}[]
\caption{Quantitative Evaluation of Class Agnostic \&   Semantic Instance Segmentation on DAVIS'17-val $AP_{50}$ score.
}
\label{tab:quantitative_MTL_DAVIS}
\centering
\resizebox{0.4\textwidth}{!}{
\begin{tabular}{|l|c|c|c|c|}
\hline
 \multirow{2}{*}{Backbone} & \multicolumn{2}{c|}{Class Agnostic} & \multicolumn{2}{c|}{Semantic} \\ \cline{2-5}
 & Mask & Box & Mask & Box \\ \hline
 ResNet101 & 23.5 & \textbf{45.4} & \textbf{42.9} & \textbf{56.2} \\
 MobileNetV2 & \textbf{28.2} & 45.3 & \textbf{42.9} & 55.6 \\ \hline
\end{tabular}
}
\vspace{-0.5cm}
\end{table}

\begin{figure*}[t!]
\captionsetup[subfigure]{labelformat=empty}
\centering




\begin{subfigure}{0.25\textwidth}
    \includegraphics[width=\textwidth]{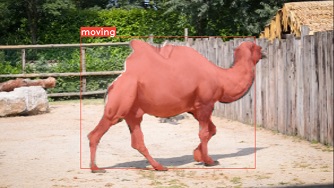}
\end{subfigure}%
\begin{subfigure}{.25\textwidth}
    \includegraphics[width=\textwidth]{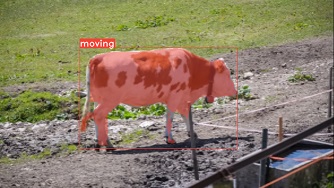}
\end{subfigure}%
\begin{subfigure}{.25\textwidth}
    \includegraphics[width=\textwidth]{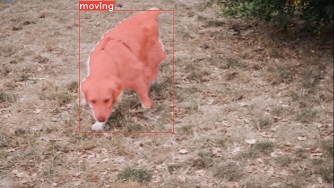}
\end{subfigure}%
\begin{subfigure}{.25\textwidth}
    \includegraphics[width=\textwidth]{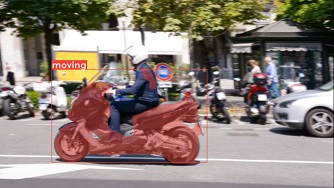}
\end{subfigure}

\quad
    \vspace{-0.1cm}
    \caption{Class-agnostic instance masks for moving objects in Multitask Model.}
    \label{fig:qualitativeEval_MTL}
\end{figure*}

Table \ref{tab:quantitative_MTL} illustrates our results for the multi-task semantic and class agnostic instance segmentation. We compare against the baseline models trained only for one task either semantic or class agnostic instance segmentation on InstanceMotSeg. Generally the multitask model performs on par with the class agnostic single task baseline, but with the added benefit of jointly predicting both semantic and class agnostic masks. Although we notice a degradation in the semantic instance segmentation performance from the baseline model to the multitask model due to much smaller size of instance segmentation dataset, we think this can be remedied by learning a re-weighting for the losses or by training the class agnostic head while freezing the weights for the rest of the model. 
Nonetheless the scope of the current work is to show the efficiency of sharing protonet among these two tasks where our multi-task model runs at 34 fps while the class agnostic baseline runs at 39 fps. Thus without large degradation in computational performance we are able to predict both semantic and class agnostic segmentation. We further validate and analyse the shared protonet in the following section. In order to assess the power of class agnostic segmentation on more general moving objects that are outside the labels within InstanceMotSeg we report results on alternate training the model between DAVIS'17 and InstanceMotSeg in Table~\ref{tab:quantitative_MTL_DAVIS}. Figure~\ref{fig:qualitativeEval_MTL} further shows the multi-task model inferring semantic labels on KITTI, while still being able to segment unknown moving objects that exist in DAVIS dataset. The multi-task model has the advantage of sharing the encoder for both the tasks and reduces computational cost drastically as encoder is the most intensive part. Finally, we evaluate our model on an unseen YouTube video with a running moose. The moose was detected using the class agnostic segmentation as shown in the qualitative results video.

\begin{figure*}[t!]
\label{fig:qualitativeEval}
\centering
\begin{subfigure}{.3\textwidth}
    \includegraphics[width=\textwidth]{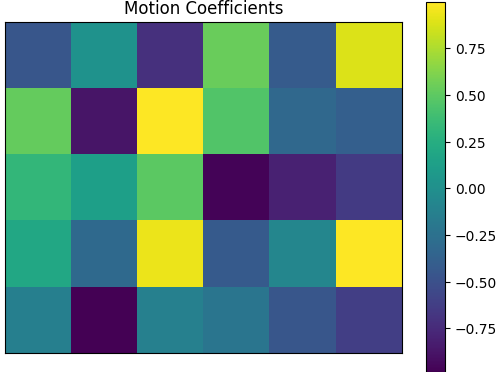}
    \caption{\textcolor{white}{(a)}}
\end{subfigure}%
\begin{subfigure}{.3\textwidth}
    \includegraphics[width=\textwidth]{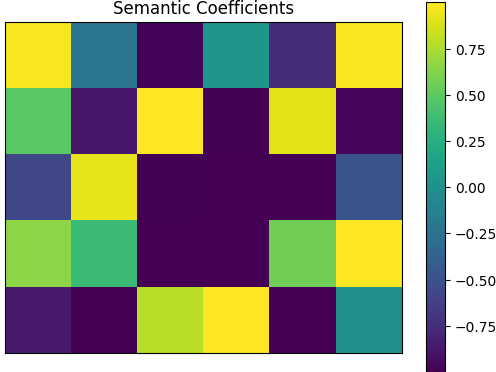}
    \caption{\textcolor{white}{(b)}}
\end{subfigure}%
\begin{subfigure}{.3\textwidth}
    \includegraphics[width=0.85\textwidth]{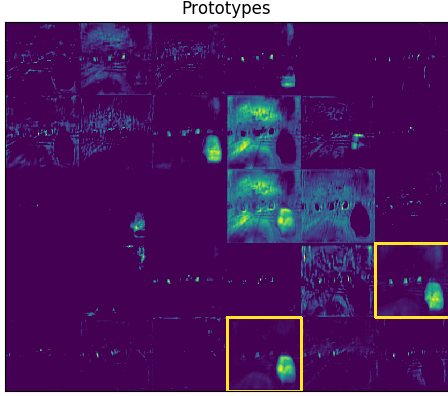}
    \caption{\textcolor{white}{(c)}}
\end{subfigure}


\begin{subfigure}{.498\textwidth}
    \includegraphics[width=\textwidth]{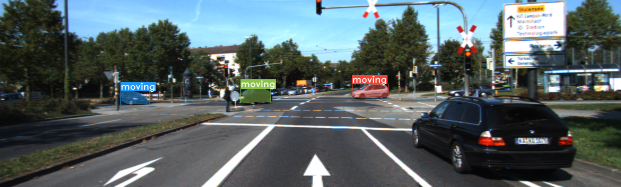}
    \caption{\textcolor{white}{(d)}}
\end{subfigure}%
\begin{subfigure}{.498\textwidth}
    \includegraphics[width=\textwidth]{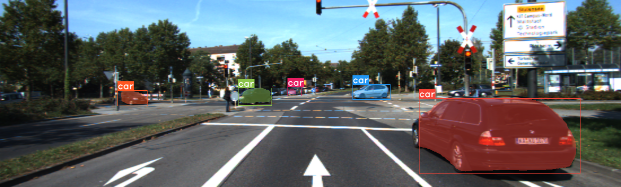}
    \caption{\textcolor{white}{(e)}}
\end{subfigure}

\caption{Prototypes and Coefficients Analysis. 
(a, b) 6 $\times$ 5 Motion and Semantic coefficients in a grid, notice that coefficients can be positive or negative. (c) 6 $\times$ 5 Prototypes grid corresponding to coefficients of 32 basis prototypes. (d, e) motion and semantic instance segmentation masks. As described in Section \ref{sec:protos_analysis}, coefficients of motion and semantic show how the model learns to linearly combine between different prototypes which are shared among the two tasks to finally get both motion and semantic instance predictions.}
\label{fig:protos}
\end{figure*}

\subsection{Prototype and Coefficients Analysis}
\label{sec:protos_analysis}
In order to better understand the model output, we perform an analysis on the common prototypes and coefficients learned for both motion and semantic instance segmentation. Figure~\ref{fig:protos} shows the output for the basis learned with a total of 32 prototypes in (c) for two different frames. The output basis are organized in a 6 $\times$ 5 grid which correspond to the 6 $\times$ 5 grid for the coefficients. We denote the different grid cells using $c^s_{i,j}$, $c^m_{i,j}$ and $p_{i,j}$ for semantic coefficients, motion coefficients and prototypes respectively. Here $i$ indicates row index starting from top to bottom and $j$ indicates column index starting from left to right. Both semantic and motion coefficients are shown in (a, b), where they can be negative or positive which can help to mask or add certain objects to the final mask. The predicted final semantic and motion instance masks for the corresponding two frames are shown last row in (d, e) which are constructed as a linear combination using these basis and coefficients. The output shows meaningful learned basis that can help in constructing the final masks. Interestingly, the comparison between motion and semantic coefficients can explain what the model is learning. 


In Figure~\ref{fig:protos}, we aim to visualize the prototypes and the coefficients to interpret our model. Images with semantic and motion instance segmentation are shown along with the generated prototypes and the corresponding weighting coefficients.
$p_{4,6}$ (row 4, col 6) shows a single prototype where all the vehicles are highlighted including the static car on the right side and the moving cars in the background. 
The activation in the corresponding semantic coefficient is high (yellow) as all the vehicles are highlighted. It is also high in case of the motion coefficients as most of these vehicles are moving. On the other hand, $p_{5,4}$ shows only the static vehicle. Thus, there is a high activation in the semantic coefficients and a low activation in the motion coefficients as the vehicle is static. If we take these two prototypes and combine them linearly using the motion coefficients, the static car will be suppressed in the motion output, but it will remain in the semantic output using the semantic coefficients. Thus it appears that the shared prototypes is sufficient to jointly handle the two tasks. Another interesting observation is that some of the prototypes are not significantly useful in highlighting an object. This indicates that we can use lesser number of prototypes to further improve the efficiency.  YOLACT~\cite{bolya2019yolact} proposed the use of 32 prototypes for handling 80 classes setting in MS-COCO. Our problem setting is much simpler with 8 classes for semantic instance segmentation.

%% file: content/conc.tex
In this paper, we developed an instance motion segmentation model for autonomous driving. As data is the bottleneck, we created InstanceMotSeg dataset by extending our previous KITTI-MoSeg datset. We provide four additional classes to improve diversity of moving objects instead of just vehicles.
The proposed instance motion segmentation model is intended to be used for class agnostic object segmentation using motion cues. 
Our model using efficient MobileNetV2 encoder is able to run at 39 fps on a Titan Xp GPU. We also extended it to a multi-task learning model which jointly performs semantic and class agnostic instance segmentation and runs at 34 fps. 
We proposed a novel design for multi-task learning based on sharing encoders and prototype generation network while learning separate coefficients for constructing the final masks. We perform exhaustive ablation study and show promising results and outperform previous motion segmentation methods. Our proposed model is a simple baseline and we hope it encourages further research to incorporate geometric constraints as inductive bias.